\pdfoutput=1
\documentclass[11pt]{article}

\usepackage{ACL2023}
\usepackage{times}
\usepackage{latexsym}

\usepackage[T1]{fontenc}
\usepackage[utf8]{inputenc}

\usepackage{microtype}

\usepackage{inconsolata}

\usepackage{graphicx}
\usepackage{enumitem}
\usepackage{url}
\usepackage{multirow}
\usepackage{amsmath}
\usepackage{booktabs}

\title{Mention Attention for Pronoun Translation}

\author{Gongbo Tang \\
  School of Information Science, \\ Beijing Language and Culture University, \\ Beijing, China\\
  \texttt{gongbo.tang@blcu.edu.cn} \\\And
  Christian Hardmeier \\
  Department of Computer Science \\ IT University of Copenhagen \\ Copenhagen, Denmark\\
  \texttt{chrha@itu.dk } \\}

\begin{document}
\maketitle
\begin{abstract}

Most pronouns are referring expressions, computers need to resolve what do the pronouns refer to, and there are divergences on pronoun usage across languages. Thus, dealing with these divergences and translating pronouns is a challenge in machine translation. Mentions are referring candidates of pronouns and have closer relations with pronouns compared to general tokens. We assume that extracting additional mention features can help pronoun translation. Therefore, we introduce an additional mention attention module in the decoder to pay extra attention to source mentions but not non-mention tokens. Our mention attention module not only extracts features from source mentions, but also considers target-side context which benefits pronoun translation. In addition, we also introduce two mention classifiers to train models to recognize mentions, whose outputs guide the mention attention. We conduct experiments on the WMT17 English--German translation task, and evaluate our models on general translation and pronoun translation, using BLEU, APT, and contrastive evaluation metrics. Our proposed model outperforms the baseline Transformer model in terms of APT and BLEU scores, this confirms our hypothesis that we can improve pronoun translation by paying additional attention to source mentions, and shows that our introduced additional modules do not have negative effect on the general translation quality. 
\end{abstract}

\section{Introduction}

Pronouns have different functions and their use varies across languages. They can pose serious problems for MT systems, which potentially causes many bad translations depending on language pairs and text types \citep{hardmeier-federico-2010-modelling,scherrer-etal-2011-la}. 
Neural machine translation (NMT) has boosted machine translation significantly in recent years \cite{kalchbrenner-blunsom-2013-recurrent,sutskever2014sequence,bahdanau15joint,luong-etal-2015-effective,gehring2017convolutional,vaswani2017Attention}. Yet NMT still suffers in pronoun translation, especially anaphoric pronouns \citep{muller-etal-2018-large,guillou-etal-2018-pronoun,Hardmeier2018error}. In this paper, we target to improve pronoun translation.

There are two main challenges in pronoun translation, one is identifying what the pronouns refer to in the source language, and the other one is the agreement of pronouns in gender or number in the target language. 
\citet{sorodoc-etal-2020-probing} have found that neural language models can capture grammatically features but not semantic referential information. 
NMT models are language models and can generate fluent sentences with few grammatical errors. Thus, dealing with anaphoric pronouns are more challenging, and building connections between pronouns and other mentions is a key for pronoun translation. 
\citet{yin-etal-2021-context} have shown that the Encoder-Decoder attention has very low alignment with supporting context for ambiguous translations.  
We assume that NMT models do not extract enough features from source-side mentions. Thus, we hypothesize that we can improve pronoun translation by paying additional attention to source mentions.

Now it arises another question: How to pay additional attention to mentions? 
Attention mechanism is very flexible and not all the words are paid attention to equally in NMT. It is not wise to change the attention weights over source mentions directly in the conventional Encoder-Decoder attention. One possible solution is that we can introduce another attention module in the decoder and this attention module only pays attention to source mentions which is different from paying attention to all the tokens twice. 
We apply simple masking constrains to let the mention attention focuses on mentions instead of using sparse attention \citep{Martins2016FromST}. 

To let models have the ability to recognize mentions, we also introduce two mention classifiers, one is in the source side and the other one is in the target-side. We jointly optimize the translation loss and the mention classification loss. The mention tags are generated by existing tools during training, and the mention classifiers predict mention tags by themselves during inference. These mention tags can guide the mention attention.  In addition, the inputs to our mention attention module are decoder hidden states which are aware of target-side context, this potentially benefits pronoun translation as well, because human translators consider target-side translations as well. 

We conduct experiments on the WMT 2017 English--German translation task, our proposed method achieves slightly better BLEU scores compared to the baseline, a conventional Transformer NMT model. 
As BLEU metric \citep{papineni-2002-machine} can not indicate the ability to translate pronouns, we use the accuracy of pronoun translation (APT) metric \citep{miculicich-werlen-popescu-belis-2017-validation} and contrastive evaluation to evaluate our models on pronoun translation. 
Our experimental results show that our proposed method achieves higher APT scores, especially for ambiguous pronouns. 
These results confirms that we can improve pronoun translation via paying additional attention to source mentions.

\section{Methodology}

\subsection{Models}
\begin{figure}[htbp]
\centering
        \includegraphics[totalheight=6.4cm]{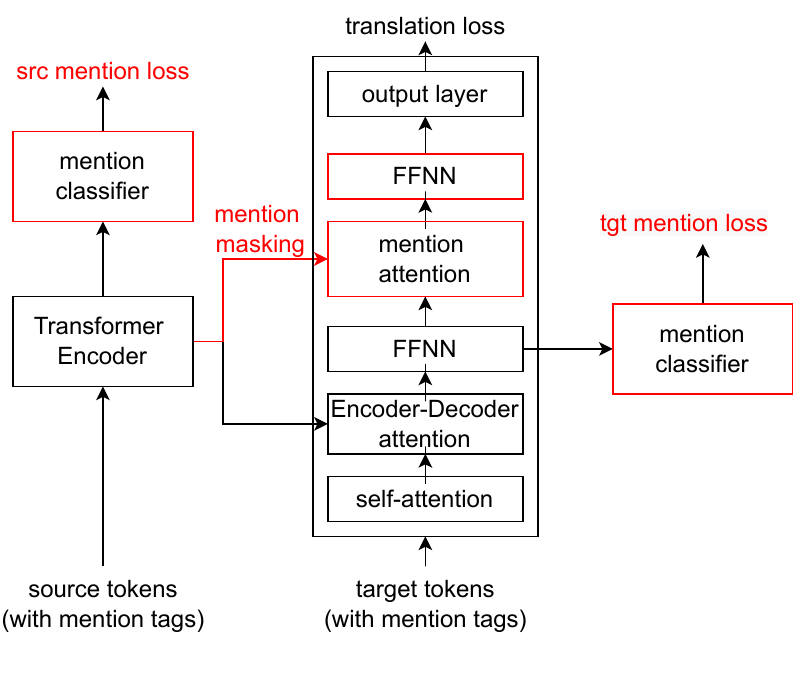}
    \caption{Illustration of model architecture with mention attention module and two mention classifiers. Red blocks and arrows are newly introduced modules compared to Transformer NMT models. }
    \label{fig:model-arch}
\end{figure}

Our model is built upon conventional Transformer NMT architecture, which is illustrated in Figure~\ref{fig:model-arch}. Compared to the base architecture, we introduce a mention attention layer in the decoder, on the top of the original feedforward neural network (FFNN) layer, to extract additional features from source-side mentions. Following the base architecture, we add another FFNN layer on the top of mention attention module, before the output layer, which can further extract features. We also introduce two mention classifiers in the encoder and the decoder, to let the model learn to distinguish mentions and other tokens. Each classifier is a FFNN, where the inputs are hidden states from the encoder or the decoder, and the outputs are mention or not. During training, we feed tokens with mention tags to the model, and let the model learn to predict mentions and translate jointly. During inference, the model itself predicts mention tags for each token to guide the mention attention to extract source features.

\subsection{Mention Attention} 

The mention attention module is used for extracting additional features from source mentions. It has the same architecture of Encoder-Decoder attention. However, the mention attention does not extract features from all the source tokens but only mention tokens. We implement it through mention masking, which is affected by mention tags, either directly from input or predicted by the mention classifiers. The input to the mention attention is the output of the first FFNN which has considered both source- and target-side context. This helps our mention attention better extract additional source-side features.

\subsection{Evaluation}
We apply BLEU and APT metrics to evaluate the performance of the Transformer model and our proposed model, for general translation and pronoun translation. In addition, we also conduct contrastive evaluation to test their abilities to recognize correct and incorrect pronoun translations. 

\textbf{APT} Since ambiguous pronouns are more challenging, we compute APT scores on both general pronouns and ambiguous pronouns. Here we focus on ``it'' and ``they'' following \citet{guillou-etal-2016-findings,loaiciga-etal-2017-findings}. 

\textbf{Contrastive Evaluation} 
Evaluation Contrastive evaluation tests the sensitivity of NMT models to specific translation errors. In contrastive evaluation sets, human reference translations are paired with one or more contrastive variants, where a specific type of error is introduced automatically. The evaluation procedure then exploits the fact that NMT models are conditional language models. If a model assigns a higher score to the correct target sentence than to a contrastive variant that contains an error, we consider it a correct decision. The accuracy of a model on such a test set is simply the percentage of cases where the correct target sentence is scored higher than all contrastive variants.

Our contrastive evaluation is based on ContraPro \citep{muller-etal-2018-large} which is a test suite of contrastive translations for pronoun translation. More specifically, it focuses on translating English pronoun “it” into German translations. The translation of “it” depends on the grammatical gender of the corresponding antecedent. Thus, the distance between the pronoun and its antecedent is crucial for pronoun translation. ContraPro can be split into several groups depending on such distance. Distance 0 means that the pronoun and the antecedent are in the same sentence. Note that although ContraPro provides inter-sentential context for each source sentence, we do not consider the context during evaluation, as the models in our experiments are not context-aware.

\begin{table*}[ht]
\centering
\scalebox{0.84}{
\begin{tabular}{ccccccccccc}
\toprule
\multirow{3}{*}{Model}&\multirow{3}{*}{BLEU} & & \multicolumn{2}{c}{APT}&&\multicolumn{5}{c}{Accuracy}\\
& &&\multirow{2}{*}{Pronouns}& \multirow{2}{*}{Ambiguous Pronouns} & &\multirow{2}{*}{Overall}&&\multicolumn{3}{c}{Distance}  \\
& &&&&&&& 0& 1 & >1\\
\midrule
Transformer & 28.0 &&60.1&50.4 & &48.7&& 76.2& 38.5 & 51.3\\
Our Model   & 28.2 &&61.2&52.2 & &48.7&& 74.3& 38.5 & 51.8 \\
\bottomrule
\end{tabular}
}
\caption{\label{table:res} Experimental results on English--German, including BLEU scores, APT scores, and accuracy. } 
\end{table*}

\section{Experiments}

We train models based on Transformer and our proposed model, and evaluate their general translation quality and the performance of pronoun translation. 
Our model code is based on public available code from \citet{Xie2022EndtoendEN}, which is based on the fairseq toolkit \citep{ott-etal-2019-fairseq}. The baseline model is trained with the fairseq toolkit as well. 
Our experiments are based on the WMT17 English--German news translation task. 

\subsection{Data}

The WMT17 English--German data set\footnote{http://www.statmt.org/wmt17/translation-task.html} has 5.8M, 3k, and 6.7k training, development, and test sentence pairs, respectively. 
For contrastive evaluation, ContrPro has 12,000 sets of contrastive examples. 

We follow \citet{tang-etal-2018-self} to process the data. We learn a joint BPE model with 32,000 subword units.  
We apply SpaCy to generate mention tags. Specifically, we first utilize spaCy to generate the simple universal POS tags of tokens, then we consider tokens belonging to the following types \{NOUN, PRON, PROPN, SYM, NUM\} as mentions and transform these tags to ``mention'', the tags of remaining tokens are changed to ``None''. 
Since all the tokens have turned into subwords after processing, we conduct a mapping between mention tags and subwords, i.e., all the subwords of a token share the same mention tag.

\subsection{Settings}
All the models are based on \textit{transformer\_base} configuration which has 6 encoder layers and 6 decoder layers. The embedding size, FFNN dimension size are 512 and 2048. The dropout rate is 0.1. Self-attention modules have 8 attention heads. The token batch size is set to 4000. We apply Adam optimizer with \text{invert\_sqrt}  learning rate schedule, and the we set the initial learning rate to 5e-4. To keep the model serving for translation and following the loss values, we set the translation loss, source-side mention loss and target-side mention loss ratio to 10:1:1. We set the maximum training epoch as 20, and select the checkpoint achieving the best perplexity score on the development set for evaluation. Each model is trained on a single GTX 1080Ti GPU. Note that our model is initialized by the baseline Transformer model.

\subsection{Results and Analysis}

Table~\ref{table:res} shows all the results of our experiments. We can see that our model achieves a slightly better BLEU score compared to the baseline Transformer model. This means that our method does not affect the general translation quality. 

Regarding the APT scores, our method beats the baseline model and achieves 0.9\% improvement on pronoun translation. If we only pay attention to ambiguous pronouns, which are more challenging, the APT score gets increased by 1.8\% which is more significant than the improvement on all the pronouns. This indicates that our mention attention module can extract additional context features for disambiguation.

For the contrastive evaluation, our model gets 48.5\% accuracy on the entire test set, which is slightly lower that of the baseline model (48.7\%). 
To further check the performance of these models, we also report the accuracy based on distance between pronouns and their antecedents. For pronouns that their corresponding antecedents appear in the same sentence, i.e., distance is 0, our model is inferior to the baseline model. One possible explanation is that the target-side features fed into the mention attention could be misleading. This need to be further tested. 
However, when the distance is longer than 1, our model performs slightly better. As both two models are not context-aware during training and inference, it is interesting that our model can achieve better performance when the distance is longer than 1. 

\citet{sennrich-2017-grammatical,muller-etal-2018-large} have emphasized that contrastive evaluation is not evaluating pronoun translation directly, but evaluating the ability to recognize correct and incorrect pronoun translations given source and target context. We need to conduct more experiments to get conclusive findings.  

\section{Conclusions and Future Work}

IIn this paper, we introduce a mention attention module, which can extract additional features from source mentions, to help pronoun translation. We also introduce two mention classifiers to let the model to recognize mentions, which can guide the mention attention. Our experimental results shows that our proposed model achieves better APT scores, especially for ambiguous pronouns which are more sensitive to source side context. This can confirm our hypothesis that we can improve pronoun translation by paying additional attention to source mentions. Moreover, our model achieves better BLEU scores on the WMT newstest2017 data set compared to the baseline, which means that the improvement on pronoun translation is not at the cost of affecting general translation quality. 

However, our findings from the contrastive evaluation on ContraPro do not accord with the findings based on the APT metric. On the one hand, we need conduct more experiments on more language pairs to get conclusive findings; On the other hand, we should be careful on pronoun evaluation and apply several evaluation metrics to get more convincing results.

\section*{Acknowledgements} % (fold)
This project is mainly funded by the Swedish Research Council (grant 2017-930), GT is also supported by Science Foundation of Beijing Language and Culture University (supported by “the Fundamental Research Funds for the Central Universities”) (22YBB36).

\bibliography{custom}
\bibliographystyle{acl_natbib}

\end{document}